# HeRo: RoBERTa and Longformer Hebrew Language Models

Vitaly Shalumov* 
vitaly.shalumov@gmail.com

Harel Haskey*
harelnh@gmail.com

April 18, 2023

## Abstract

In this paper, we fill in an existing gap in resources available to the Hebrew NLP community by providing it with the largest so far pre-train dataset **HeDC4**, a state-of-the-art pre-trained language model **HeRo** for standard length inputs and an efficient transformer **LongHeRo** for long input sequences. The HeRo model was evaluated on the sentiment analysis, the named entity recognition, and the question answering tasks while the LongHeRo model was evaluated on the document classification task with a dataset composed of long documents. Both HeRo and LongHeRo presented state-of-the-art performance. The dataset and model checkpoints used in this work are publicly available[1].

## 1 Introduction

Hebrew NLP research has gained a huge boost with the introduction of a multi-lingual encoder-only transformer model mBERT, based on the BERT architecture (Devlin et al., 2018). Following mBERT, several other models were released, among them are the monolingual models HeBERT (Chriqui and Yahav, 2021) and AlephBERT (Seker et al., 2021). The AlephBERT model is considered state-of-the-art (SOTA) for Hebrew pre-trained language models (PLMs). Shortly before the release of our paper, a new model named AlephBERTGimmel (Guetta et al., 2022) was released with a two and a half times larger vocabulary size than the AlephBERT model. This model has passed AlephBERT on several downstream tasks and we will address this model (the "base" version which exhibited the best results) if applicable.

---

*Equal contribution.
[1]The resources are available at the huggingface hub under the names **HeNLP/HeRo, HeNLP/LongHeRo, and HeNLP/HeDC4.**

While those models provided a means for performing NLP tasks for standard length inputs (up to 512 tokens), the problem of dealing with long input sequences (larger than 512 tokens) was left untouched. Theoretically, there is no limitation of using the aforementioned models for long documents. However, the dense attention approach adopted in those models suffers from the fact that the computational complexity of a vanilla transformer architecture grows quadratically with the length of the input.

In recent years, efficient transformers which leverage the sparse attention approach were suggested, aiming at dealing with large input sequences. Among them are the Longformer (Beltagy et al., 2020) and the BigBird (Zaheer et al., 2020) models. Both models initiated their weights using the pre-trained RoBERTa model (Liu et al., 2019). The Longformer model for the English language was trained on the Books dataset (Zhu et al., 2015) with a combination of Wikipedia, Realnews (Zellers et al., 2019) with larger than 1200 tokens inputs and a third of the Stories dataset (Trinh and Le, 2018). Similarly, BigBird model for the English language used CC-News (Guu et al., 2020) combined with Books and Stories. In this paper, the Longformer architecture was chosen.

In the scope of pre-training data, the previous SOTA model (AlephBERT) used mainly the OSCAR 2019 (Suárez et al., 2020) line-based multi-lingual pre-train dataset. In 2021, following the paper on the sequence to sequence multi-lingual model mT5 (Xue et al., 2020), the mC4 corpora which includes the Hebrew language was released. In 2022, the OSCAR 22.01 (Abadji et al., 2022) multi-lingual document-based dataset was released which is more suitable for pre-training large language models then the 2019 version. One of the main changes from the 2019 version is the utilization of line filtering intended to limit the integrity destruction of the documents, keeping contiguous lines and making documents more human readable and exploitable as documents.

In terms of quality of the pre-train dataset, the authors of (Lee et al., 2021) highlighted the importance of deduplication and suggested two approaches for deduplication - exact and approximate. In this paper, we adopt the Approximate Matching deduplication method using MinHash which is more suitable for documents. (Broder, 1997).

We followed the methodology introduced in (Beltagy et al., 2020) and trained a Longformer model for the Hebrew language from a RoBERTa checkpoint. No RoBERTa model for the Hebrew language existed at the time of writing this paper. In addition, we wanted to test the effect of the HeDC4 dataset on the performance of a standard input language model, thus we trained a SOTA Hebrew language model for standard length inputs using the RoBERTa architecture. The models for standard and long inputs are named HeRo and LongHeRo respectively.

This work introduces the following contributions:

- Synthesizing a thoroughly cleaned and approximately deduplicated dataset **HeDC4** for unsupervised learning, two and a half times larger than was previously used in the literature.

- Achieving SOTA results on several Hebrew NLP tasks for a standard input length of 512 tokens using the **HeRo** model.

- Introducing and evaluating the **LongHeRo** language model for long documents in Hebrew.

- Publicly releasing the curated dataset HeDC4, the HeRo Hebrew language model for standard length inputs and the LongHeRo Hebrew language model for long documents.

## 2 Pre-train

### 2.1 Pre-train Dataset

Both HeBERT and AlephBERT utilized the deduplicated split of the OSCAR 2019 dataset as the main source of pre-train data. Both also included the Hebrew Wikipedia. In addition, the authors of AlephBERT included a twitter-based dataset.

Due to the fact that the main objective of this research was to train a model for long input sequences, we aimed at curating a dataset that has a large amount of long documents (larger than 512 tokens). Thus, we combined the mC4 and the OSCAR 22.01 datasets for pre-training. The mC4 and OSCAR 22.01 datasets are not deduplicated, thus we used the Approximate Matching deduplication to filter duplicate documents within each dataset and between the two datasets.

Following the guidelines from (Sarti and Nissim, 2022), we also added a cleaning stage to our dataset pre-processing procedure with the exception of retaining long documents. After deduplication and cleaning, we achieved a two and a half times larger dataset than the one used in AlephBERT training, making it the most extensive publicly available Hebrew corpus to this date. The dataset is composed of about 30% long documents. Tab. 1 presents the details of the curated dataset, which we name Hebrew Deduplicated and Cleaned Common Crawl Corpus (HeDC4).

### 2.2 Pre-training HeRo

We began by constructing a BPE tokenizer (Sennrich et al., 2015) with a vocabulary size of 50,265 tokens using the HeDC4 dataset. The model was trained in two stages: in the first stage the model was trained on four NVIDIA GeForce GTX 1080 Ti 12GB GPUs for 35 days. The checkpoint was then saved for LongHeRo initialization. The second stage of training was identical to the first. Each stage started with the learning rate of 1e-4 with linear decay and the AdamW optimizer (Loshchilov and Hutter, 2017).

### 2.3 Pre-training LongHeRo

We initiated the LongHeRo model using the weights taken from the HeRo model at the end of the first training stage. The LongHeRo model was then trained for 2 epochs using the learning rate of 3e-5 with linear decay and the AdamW optimizer. The model was trained on a NVIDIA Quadro RTX 6000 48GB GPU for 33 days.

Table 1: The size of the pre-train dataset after each pre-processing stage. Previous largest corpora has been reported in (Seker et al., 2021) as 17.9GB file size and 1.9B words.

| Corpus | File Size | Documents | Words |
| --- | --- | --- | --- |
| Hebrew mC4 | 74 GB | 12.3M | 7.2B |
| Hebrew OSCAR 22.01 | 30.3GB | 3.1M | 2.2B |
| Hebrew deduplicated mC4 and OSCAR 22.01 | 57.1GB | 9.5M | 5.6B |
| **HeDC4** (deduplicated and cleaned) | **47.5GB** | **8.5M** | **4.7B** |

## 3 Downstream Tasks

To facilitate comparability with previous works, we retain the terminology *morph-based* evaluation which refers to pre-processing the input by using morphological segmentation. The *token-based* phrase references evaluation without morphological segmentation. The performance of the models in this paper is reported for the token-based segmentation.

### 3.1 HeRo Downstream Tasks

We focus on comparing our model to the current SOTA as presented in (Seker et al., 2021). For this purpose, we evaluate our model on three NLP tasks: sentiment analysis (SA), named entity recognition (NER), and question answering (QA).

#### 3.1.1 Sentiment Analysis

The HeRo model was evaluated on the sentiment analysis dataset presented in (Amram et al., 2018) which is based on 12K social media comments. The comments were written in response to official Facebook posts posted by the former Israeli president, Mr. Reuven Rivlin, between June and August, 2014. The comments were annotated with the following labels - supportive (positive), criticizing (negative), or off-topic (neutral). As stated in (Seker et al., 2021), the original variant of the dataset had a significant data leakage between the splits, with duplicates in the data samples. Thus, the task was evaluated on the deduplicated dataset[2] which contains 8,465 samples.

[2]The dataset was taken from https://github.com/omilab/Neural-Sentiment-Analyzer-for-Modern-Hebrew

The dataset will be henceforth referenced as SMCD (social media comments deduplicated).

#### 3.1.2 Named Entity Recognition

Two NER datasets were established as the benchmark for the Hebrew NER task. The first is the BMC corpus (Mordecai and Elhadad, 2005) which contains 3,294 sentences (4,600 entities) and has seven different entity categories.

The second is the Named Entities and MOrphology (NEMO) corpus presented in (Bareket and Tsarfaty, 2021). The NEMO corpus has nine categories and contains 6,220 sentences (7,713 entities).

#### 3.1.3 Question Answering

We evaluate our model on the ParaShoot dataset (Keren and Levy, 2021) composed of 3,038 annotated examples with SQuAD format.

### 3.2 LongHeRo Classification Task

While there are numeral benchmarks for the evaluation of Hebrew language models with standard length inputs, no labeled data existed for an evaluation of a long input Hebrew language model. Thus, we turned to creating a labeled dataset for specifically that purpose.

We chose to translate the Hyperpartisan news detection dataset (Kiesel et al., 2019) on which the English version of both RoBERTa and Longformer models were evaluated. This choice enabled us to compare not only the performance of HeRo to that of LongHeRo, but also served as a sanity check in comparing the Hebrew based models to their English counterparts. We chose machine translation under the assumption that the generated translation is of sufficient quality for the preservation of the document class.

As stated in (Beltagy et al., 2020), documents in the Hyperpartisan dataset are rel-atively long and the dataset is small with only 645 documents making it a good test for LongHeRo's ability to adapt to limited data.

Translating with Helsinki and NLLB models introduced in (Tiedemann and Thottingal, 2020) and (Costa-jussà et al., 2022) respectively, proved to be difficult with an off-the-shelf inference approach. While most of the translated text appeared coherent, some phrases caused the model to repeat a certain translated sequence several times. Thus, the Google Translate tool[3] was eventually chosen to translate the Hyperpartisan dataset from English to Hebrew. We validated manually a subset of the dataset for quality control.

## 4 Core Results

### 4.1 HeRo Results

We fine-tuned the model for 25 epochs with the default Huggingface (Wolf et al., 2020) parameters. The results for the sentiment analysis task are given in Tab. 2.

Table 2: Token-based SA accuracy results on the SMCD corpus. Previous SOTA has been reported by the AlephBERTGimmel model (Guetta et al., 2022).

| Model | SMCD |
| --- | --- |
| AlephBERT | 89.02 |
| Previous SOTA (AlephBERTGimmel) | 89.51 |
| HeRo | **89.56** |

To the best of our knowledge, the NER dataset on which the AlephBERTGimmel model was evaluated was not released at the time of writing this paper, thus we chose to evaluate this model on the datasets NEMO and BMC using the released checkpoint. The results for the NER task are given in Tab. 3.

Table 3: Token-based NER F1 results on NEMO and BMC datasets. Previous SOTA on both corpora has been reported by the AlephBERT model (Seker et al., 2021).

| Model | NEMO | BMC |
| --- | --- | --- |
| Previous SOTA on BMC (AlephBERT) | 84.91 | 91.12 |
| Previous SOTA on NEMO (AlephBERTGimmel) | 86.76 | 91.11 |
| HeRo | **86.79** | **93.56** |

The HeRo model presents similar performance to the AlephBERTGimmel model on SA and NER (NEMO) and outperforms it on NER (BMC). It is interesting to note that this level of performance is accomplished with a much smaller model (AlephBERTGimmel has 50% more parameters then than HeRo due to the increased vocabulary: AlephBERTGimmel - 184M parameters; HeRo and AlephBERT: roughly the same size of 125M parameters). This fact is attributed to the two and a half times larger corpus (HeDC4) introduced in this paper.

The results for the QA task are given in Tab. 4. On the QA task the HeRo model outperforms both the AlephBERT model and the multi-lingual mBERT model.

Table 4: QA F1 results on the ParaShoot corpus. The results are compared to other models based on (Keren and Levy, 2021). Previous SOTA has been reported by the mBERT model (Devlin et al., 2018).

| Model | F1 | Exact Match |
| --- | --- | --- |
| Previous SOTA (mBERT) | 56.1 | 32.0 |
| AlephBERT | 49.6 | 26.0 |
| HeRo | **58.2** | **34.8** |

With respect to dataset diversity, we've experimented with adding several gigabytes of tweeter-based documents to the HeDC4 dataset. This addition had little effect on performance and was thus discarded.

### 4.2 LongHeRo Classification

We fine-tuned the HeRo and LongHeRo models for 15 epochs with the default Huggingface (Wolf et al., 2020) parameters. The results represent a mean of F1 scores on three random 80/20 splits. The results for the classification task are given in Tab. 5.

Table 5: Token-based classification F1 results on the Hyperpartisan dataset. The RoBERTa and Longformer for the English language were taken from "roberta-base" and "allenai/longformer-base-4096" checkpoints hosted on the Huggingface hub (Wolf et al., 2020), respectively.

| Model | Hyperpartisan (En) | Hyperpartisan (He) |
| --- | --- | --- |
| RoBERTa (English) | 87.3 | — |
| HeRo | — | 86.9 |
| Longformer (English) | 90.1 | — |
| LongHeRo | — | 89.1 |

Tab. 5 demonstrates the compatibility of the Hebrew models to their English counter-parts and the improvement of the LongHeRo model over the standard length input model HeRo.

---
[3] https://translate.google.com/

The training process exhibited continuing improvement in both the pre-training loss and the performance on the downstream tasks. This suggests that further pre-training of the released models in conjunction with the released dataset will yield even higher performance. Thus, we encourage the users to continue pre-training the released models on the released dataset prior to fine-tuning on downstream tasks.

## 5 Conclusions

In this paper, we presented several additional resources for the Hebrew NLP community. We introduced a new pre-train dataset HeDC4 for the Hebrew language, which is two and a half times larger than the one previously used in the literature. This dataset was used to train a new language model HeRo, based on the RoBERTa architecture. A mid-training checkpoint of the HeRo model was then used to train the first Hebrew language model for long documents LongHeRo, based on the Longformer architecture.

The models were evaluated on several downstream tasks and compared to the results presented in the literature. The HeRo model presented state-of-the-art results on the sentiment analysis, the named entity recognition tasks, and the question answering task. The LongHeRo model was evaluated on the document classification task using a translated dataset. The HeRo model was evaluated as a baseline. The performance of both models were found compatible with their English counterparts. For future work we aim to further investigate the performance of the LongHeRo model by evaluating it on other tasks involving long documents such as document similarity.

## 6 Acknowledgements

We are grateful to Tal Geva and Amir David Nissan Cohen for technical discussions during the project.

## A  Near-Duplicate Example

The example in Tab. 6 demonstrates the near-duplicate document which was present in the dataset prior to deduplication. The similarity between documents is colored in blue. It is clear that the only difference between the documents is in the short starting sentence.

Table 6: An example of an identified near-duplicate document.

| Example - Doc.Id 135 | Near-duplicate example - Doc.Id 89087 |
| --- | --- |
| 'קריינות: בגן הזה הילדים מדברים בעברית ובערבית וחוגגים בו חגים של שלושת הדתות. הוא הוקם ביוזמת הורים יהודים וערבים ביפו, שביקשו כי ילדיהם ילמדו במסגרות משותפות, ללא מחיצות והפרדה.לאחר שנתיים של פעילות, עלה הצורך הטבעי במסגרת המשכית - הקמת בית ספר יהודי-ערבי. כמו עם הקמת הגנים, גם כאן, נתקלו ההורים בהתנגדות העירייה.אופיר קהת: בעצם מה שאנחנו מבקשים זה דבר מאד פשוט.  אנחנו רוצים להמשיך עם ההצלחה שלנו של גני הילדים הדו לשוניים ולפתוח בית ספר צומח שימשש את השאיפות שלנו לדו קיום.מרב קליין אשר: אני חושבת שבית ספר דו לשוני הוא הרבה יותר מאשר רק ללמוד את השפה. יש משהו ביכולת של ילדים לא להיות גזעניים, שזה דבר נערץ ומקסים ושובה לב ואני חושבת שבאמת היכולת ללמוד את השפות אחד של השני מתוך מקום של כבוד והערכה ושוויון אמיתי.קריינות: העירייה מצידה הציעה להורים להשתלב בבית ספר ממלכתי קיים במקום להקים בית ספר דו לשוני, בשלב זה פתרון זה לא נתקבל על ידי ההורים בחיוב.אופיר קהת: הסיבה שאנחנו לא מעוניינים להירשם לבית הספר החלופי, שאותו העירייה מציעה, מדובר בבית ספר במסלול עברי שבו המנהל לא מאפשר לערבים לדבר בערבית.סופי קאסם: אנחנו רוצים לחנך את הילדים שלנו ללמוד גם את הערכים של היהודים וגם את הערכים שאנחנו לא רוצים לאבד את השפה שלנו, את הדת שלנו.מרב קליין אשר: המערכת עצמה היא מערכת שמעדיפה את העברית, שמעדיפה את היהודי, שמעדיפה את הסמלים היהודיים שנותנת כבוד לחגים היהודיים. ואני חושבת שבאמת כילד ערבי ביפו לגדול בלי שהגן שלך והבית ספר שלך והמורה שלך מכבדת את הזהות שלך את החגים שלך את השפה שלך, זה חינוך לאפליה זה חינוך לגזענות.סופי קאסם: כולנו עובדים בשביל הילדים שלנו כולנו רוצים לחנך את הילדים שלנו. לכולנו יש אותן מטרות לחנך את הילדים שלנו לעולם יותר טוב והבית ספר הזה עונה על הדרישות שלנו.קריינות: כעבור כמה ימים הגיעו הצדדים לפשרה. העירייה התחייבה כי בכיתות הדו לשוניות בבית הספר הממלכתי יהודי ילמדו שתי מורות ערבייה ויהודיה מו כן החגים יכלול את כל החגים של היהודים והערבים.' | 'הכתבה הופקה על-ידי: הטלווזיה החברתית. קריינות: בגן הזה הילדים מדברים בעברית ובערבית וחוגגים בו חגים של שלושת הדתות. הוא הוקם ביוזמת הורים יהודים וערבים ביפו, שביקשו כי ילדיהם ילמדו במסגרות משותפות, ללא מחיצות והפרדה.לאחר שנתיים של פעילות, עלה הצורך הטבעי במסגרת המשכית - הקמת בית ספר יהודי-ערבי. כמו עם הקמת הגנים, גם כאן, נתקלו ההורים בהתנגדות העירייה.אופיר קהת: בעצם מה שאנחנו מבקשים זה דבר מאד פשוט. אנחנו רוצים להמשיך עם ההצלחה שלנו של גני הילדים הדו לשוניים ולפתוח בית ספר צומח שימשש את השאיפות שלנו לדו קיום.מרב קליין אשר: אני חושבת שבית ספר דו לשוני הוא הרבה יותר מאשר רק ללמוד את השפה. יש משהו ביכולת של ילדים לא להיות גזעניים, שזה דבר נערץ ומקסים ושובה לב ואני חושבת שבאמת היכולת ללמוד את השפות אחד של השני מתוך מקום של כבוד והערכה ושוויון אמיתי.קריינות: העירייה מצידה הציעה להורים להשתלב בבית ספר ממלכתי קיים במקום להקים בית ספר דו לשוני, בשלב זה פתרון זה לא נתקבל על ידי ההורים בחיוב.אופיר קהת: הסיבה שאנחנו לא מעוניינים להירשם לבית הספר החלופי, שאותו העירייה מציעה, מדובר בבית ספר במסלול עברי שבו המנהל לא מאפשר לערבים לדבר בערבית.סופי קאסם: אנחנו רוצים לחנך את הילדים שלנו ללמוד גם את הערכים של היהודים וגם את הערכים שאנחנו לא רוצים לאבד את השפה שלנו, את הדת שלנו.מרב קליין אשר: המערכת עצמה היא מערכת שמעדיפה את העברית, שמעדיפה את היהודי, שמעדיפה את הסמלים היהודיים שנותנת כבוד לחגים היהודיים. ואני חושבת שבאמת כילד ערבי ביפו לגדול בלי שהגן שלך והבית ספר שלך והמורה שלך מכבדת את הזהות שלך את החגים שלך את השפה שלך, זה חינוך לאפליה זה חינוך לגזענות.סופי קאסם: כולנו עובדים בשביל הילדים שלנו כולנו רוצים לחנך את הילדים שלנו. לכולנו יש אותן מטרות לחנך את הילדים שלנו לעולם יותר טוב והבית ספר הזה עונה על הדרישות שלנו.קריינות: כעבור כמה ימים הגיעו הצדדים לפשרה. העירייה התחייבה כי בכיתות הדו לשוניות בבית הספר הממלכתי יהודי ילמדו שתי מורות ערבייה ויהודיה כמו כן לוח החגים יכלול את כל החגים של היהודים והערבים.' |